\documentclass[sigconf]{acmart}
\usepackage{graphicx}
\usepackage{multirow}
\usepackage{svg}
\usepackage{bm}
\usepackage{amsmath}
\usepackage{amsfonts}
\usepackage{balance}
\AtBeginDocument{%
  }

\copyrightyear{2024}
\acmYear{2024}
\setcopyright{acmlicensed}
\acmConference[KDD '24] {Proceedings of the 30th ACM SIGKDD Conference on Knowledge Discovery and Data Mining }{August 25--29, 2024}{Barcelona, Spain.}
\acmBooktitle{Proceedings of the 30th ACM SIGKDD Conference on Knowledge Discovery and Data Mining (KDD '24), August 25--29, 2024, Barcelona, Spain}
\acmISBN{979-8-4007-0490-1/24/08}
\acmDOI{10.1145/3637528.3671855}
\begin{document}

%%
%% The "title" command has an optional parameter,
%% allowing the author to define a "short title" to be used in page headers.
\title{Generative Pretrained Hierarchical Transformer for \\ Time Series Forecasting}

%%
%% The "author" command and its associated commands are used to define
%% the authors and their affiliations.
%% Of note is the shared affiliation of the first two authors, and the
%% "authornote" and "authornotemark" commands
%% used to denote shared contribution to the research.
\author{Zhiding Liu}
\orcid{0000-0003-0994-473X}
\affiliation{%
  \institution{University of Science and Technology of China}
  \department{State Key Laboratory of Cognitive Intelligence}
  \city{Hefei}
  \state{Anhui}
  \country{China}
}
\email{zhiding@mail.ustc.edu.cn}

\author{Jiqian Yang}
\orcid{0009-0007-6421-2626}
\affiliation{%
  \institution{University of Science and Technology of China}
  \department{State Key Laboratory of Cognitive Intelligence}
  \city{Hefei}
  \state{Anhui}
  \country{China}
}
\email{yangjq@mail.ustc.edu.cn}

\author{Mingyue Cheng}
\authornote{Mingyue Cheng is the corresponding author.}
\orcid{0000-0001-9873-7681}
\affiliation{%
  \institution{University of Science and Technology of China}
  \department{State Key Laboratory of Cognitive Intelligence}
  \city{Hefei}
  \state{Anhui}
  \country{China}
}
\email{mycheng@ustc.edu.cn}

\author{Yucong Luo}
\orcid{0000-0003-0685-0834}
\affiliation{%
  \institution{University of Science and Technology of China}
  \department{State Key Laboratory of Cognitive Intelligence}
  \city{Hefei}
  \state{Anhui}
  \country{China}
}
\email{prime666@mail.ustc.edu.cn}

\author{Zhi Li}
\orcid{0000-0002-8061-7486}
\affiliation{%
  \institution{Shenzhen International Graduate School, Tsinghua University}
  \city{Shenzhen}
  \state{Guangdong}
  \country{China}
}
\email{zhilizl@sz.tsinghua.edu.cn}

%%
%% By default, the full list of authors will be used in the page
%% headers. Often, this list is too long, and will overlap
%% other information printed in the page headers. This command allows
%% the author to define a more concise list
%% of authors' names for this purpose.
\renewcommand{\shortauthors}{Zhiding Liu et al.}

%%
%% The abstract is a short summary of the work to be presented in the
%% article.

\begin{abstract}
Recent efforts have been dedicated to enhancing time series forecasting accuracy by introducing advanced network architectures and self-supervised pretraining strategies. Nevertheless, existing approaches still exhibit two critical drawbacks. Firstly, these methods often rely on a single dataset for training, limiting the model's generalizability due to the restricted scale of the training data. Secondly, the one-step generation schema is widely followed, which necessitates a customized forecasting head and overlooks the temporal dependencies in the output series, and also leads to increased training costs under different horizon length settings.

To address these issues, we propose a novel generative pretrained hierarchical transformer architecture for forecasting, named \textbf{GPHT}. There are two aspects of key designs in GPHT. On the one hand, we advocate for constructing a mixed dataset under the channel-independent assumption for pretraining our model, comprising various datasets from diverse data scenarios. This approach significantly expands the scale of training data, allowing our model to uncover commonalities in time series data and facilitating improved transfer to specific datasets. On the other hand, GPHT employs an auto-regressive forecasting approach, effectively modeling temporal dependencies in the output series. Importantly, no customized forecasting head is required, enabling \textit{a single model to forecast at arbitrary horizon settings.} We conduct sufficient experiments on eight datasets with mainstream self-supervised pretraining models and supervised models. The results demonstrated that GPHT surpasses the baseline models across various fine-tuning and zero/few-shot learning settings in the traditional long-term forecasting task, providing support for verifying the feasibility of pretraining time series large models. We make our codes publicly available\footnote{https://github.com/icantnamemyself/GPHT}.

\end{abstract}

%%
%% The code below is generated by the tool at http://dl.acm.org/ccs.cfm.
%% Please copy and paste the code instead of the example below.
%%
\begin{CCSXML}
<ccs2012>
   <concept>
       <concept_id>10002950.10003648.10003688.10003693</concept_id>
       <concept_desc>Mathematics of computing~Time series analysis</concept_desc>
       <concept_significance>500</concept_significance>
       </concept>
   <concept>
       <concept_id>10010147.10010178</concept_id>
       <concept_desc>Computing methodologies~Artificial intelligence</concept_desc>
       <concept_significance>500</concept_significance>
       </concept>
 </ccs2012>
\end{CCSXML}

\ccsdesc[500]{Mathematics of computing~Time series analysis}
\ccsdesc[500]{Computing methodologies~Artificial intelligence}

%%
%% Keywords. The author(s) should pick words that accurately describe
%% the work being presented. Separate the keywords with commas.
\keywords{Time series forecasting; deep learning; pretraining}

%% A "teaser" image appears between the author and affiliation
%% information and the body of the document, and typically spans the
%% page.
% \begin{teaserfigure}
%   \includegraphics[width=\textwidth]{sampleteaser}
%   \caption{Seattle Mariners at Spring Training, 2010.}
%   \Description{Enjoying the baseball game from the third-base
%   seats. Ichiro Suzuki preparing to bat.}
%   \label{fig:teaser}
% \end{teaserfigure}

% \received{20 February 2007}
% \received[revised]{12 March 2009}
% \received[accepted]{5 June 2009}

%%
%% This command processes the author and affiliation and title
%% information and builds the first part of the formatted document.
\maketitle

\section{Introduction}

Time series forecasting is one of the fundamental tasks in time series analysis, garnering significant attention over the past several years \cite{petropoulos2022forecasting, wen2022transformers}. Precise forecasting plays a crucial role in assisting various real-world applications, including climate report \cite{nature, nature2}, patient vital sign assessments \cite{lu2023composite}, urban computing \cite{jin2023spatio} and stock prediction\cite{jiang2023forecasting} etc. Various efforts have been devoted to this area for more accurate forecasting. Notably, deep-learning-based methods have achieved great success due to their capability to capture both temporal and cross-dimension dependencies \cite{patchtst, crossformer, san}.

On the other hand, inspired by recent significant advancements in pretraining methods in both the NLP and CV fields \cite{grill2020bootstrap, mae, devlin2018bert, gpt3}, various pretraining-based time series analysis methods have been proposed \cite{ssl_survey}. The contrastive learning technique is widely employed in the discriminative pretraining methods, where the models are expected to learn representations from constructed positive and negative pairs \cite{yue2022ts2vec, woo2021cost}. Furthermore, the incorporation of generative targets, such as masked time-series modeling, into the pretraining task has also been well-studied, with the aim of extracting general knowledge during the reconstruction process \cite{kdd_pretraining, cheng2023timemae, dong2023simmtm}. Additionally, considering the shared characteristics between time series and natural languages, some recent studies have emerged to adapt pre-trained language models into accurate forecasters through prompting or fine-tuning \cite{gruver2023large, fpt}. All these methods have achieved significant success, even competing effectively with supervised forecasting approaches. 

Despite their effectiveness, there remain significant challenges in promoting the performance of the pretrained forecasters. Firstly, the limited scale of the dataset is a critical issue. The standard practice of these methods involves pretraining on a single real-world or synthetic dataset, and evaluating its performance on this dataset or other datasets through transfer learning \cite{timegpt2, dooley2023forecastpfn}. However, the amount of training instances in a single dataset is often limited \cite{dataset}, and its inherent patterns may fail to encompass the complex scenarios of other datasets, leading to suboptimal forecasting accuracy and transferability. Secondly, nearly all forecasting approaches adhere to the one-step generation schema  \cite{informer}, which implies that the predictions for all future time steps are generated through a single forward pass with a customized forecasting head determined by a specified horizon length. The drawbacks of this paradigm are multifaceted. On the one hand, the temporal dependencies within the predicted series are inevitably overlooked, potentially leading to an inferior result. On the other hand, the tailored forecasting head hinders the generalizability of the pretrained models, as multiple models are required for different horizon length settings. 

To alleviate the above challenges, we are motivated to explore pretraining a single unified forecasting model that generalizes well across diverse data scenarios and forecasting settings with a novel generative hierarchical transformer architecture, namely \textbf{GPHT}. Firstly for the dataset construction, we extend the channel-independent assumption \cite{DLinear} into multiple data scenarios, simply mixing time series originating from various scopes as a whole without considering extra information, which provides vast characteristics such as diverse periodicities that benefit the pertaining procedure. Besides, to better capture the commonalities and specialties within the mixed dataset, we naturally introduce a novel hierarchical transformer architecture as the backbone model \cite{challu2022nhits,shabani2022scaleformer}. Furthermore, we formulate the forecasting task as a standard language modeling task with the patching technique \cite{cheng2024learning, patchtst}, which projects time series into \textit{token-level} representations, and an auto-regressive optimization function is applied in our pretraining procedure to replace the conventional one-step generating schema. Consequently, the model can well model the temporal dependencies in the forecasting horizons at a token-wise level through auto-regressive inference and can be seamlessly adapted to diverse datasets with varying horizon length settings without any modification. We compare our model's performance with state-of-the-art supervised and pretraining methods under various fine-tuning and zero/few-shot learning settings. The results demonstrate the superiority and generalizability of the proposed GPHT. 

In summary, our contributions are as follows:
\begin{itemize}
    \item We explore pretraining a single unified forecasting model that generalizes well across diverse scenarios with the pretraining dataset constructed under the channel-independent assumption, which allows for the easy creation of diverse, large-scale datasets, forming the foundation for the generalizability across data scenarios of the forecasting model.
    \item We introduce GPHT, a novel hierarchical transformer forecasting in an auto-regressive manner. This design inherently aids in modeling both the commonalities and specialties within the mixed dataset, guaranteeing universality under various forecasting settings.
    \item We conduct sufficient experiments on 8 widely used benchmark datasets, comparing our proposed GPHT with mainstream supervised and pretraining methods. The results show that our model surpasses the baseline models across various fine-tuning and zero/few-shot learning settings.
\end{itemize}

%    We demonstrate the feasibility of constructing a mixed dataset for pretraining with minimal effort, utilizing various real-world forecasting datasets under the channel-independent assumption. 
%     We extend the channel-independent assumption for constructing a pretraining dataset that mixes time series originating from various scopes as a whole. This approach allows for the easy creation of diverse, large-scale datasets, forming the foundation for the generalizability across data scenarios of the forecasting model.

\section{Related Works}
\subsection{Time Series Forecasting}
Time series forecasting is a crucial task with broad applications, garnering significant attention in recent years. Early research predominantly focuses on statistical methods such as ARIMA \cite{box1968some, zhang2003time}, which builds an auto-regressive model and forecasts in a moving average fashion. However, these methods may encounter limitations in long-term forecasting settings. The advent of deep learning has led to the development of numerous models capturing both temporal and cross-dimensional dependencies in multivariate time series, utilizing modern architectures \cite{petnehazi2019recurrent, deepar, lou2022mts, SCINet, cheng2024convtimenet}.

Transformer-based and MLP-based approaches have emerged as research hotspots due to their outstanding performance \cite{chen2022learning, liu2023itransformer, DLinear}. Beyond model architecture, several customized techniques rooted in time series analysis have been established. These include trend-seasonal decomposition \cite{wu2021autoformer}, time-frequency conversion \cite{fedformer}, series stabilization \cite{RevIN}, and patching \cite{patchtst}. The integration of these advanced studies enables contemporary forecasters to achieve unprecedented accuracy in predictions across diverse scenarios through supervised training.

\subsection{Self-supervised Pretraining in Time Series Modeling}
\subsubsection{Discriminative methods}
Contrastive learning is widely utilized in the discriminative time series modeling approaches, aiming to derive crucial representations from pre-defined positive and negative pairs. A key challenge lies in effectively constructing informative instance pairs. 

In practice, TNC \cite{tnc} takes advantage of the local smoothness of a signal’s generative process to define neighborhoods in time with stationary properties, and TS2Vec \cite{yue2022ts2vec} proposes to employ contrastive learning on both instance level and patch level in a hierarchical way for robust contextual representation learning. Moreover, TS-TCC \cite{ts-tcc} introduces a new contrastive learning task of cross-view prediction. Subsequently, CoST \cite{woo2021cost} comprises both time domain and frequency domain contrastive losses to learn discriminative trend and seasonal representations. Note that discriminative methods primarily concentrate on coarse-grained instance-level information, leading to unsatisfactory performance in forecasting tasks where fine-grained temporal features are essential.

\subsubsection{Generative methods}
Generative pretraining methods usually follow a paradigm of reconstruction. In the context of time series analysis, the objective of masked time-series modeling has been extensively explored. TST \cite{tst} pioneers the use of traditional masked modeling, aiming to predict the removed time series points based on the remaining ones. Subsequently, STEP \cite{kdd_pretraining} and PatchTST \cite{patchtst} expand on this concept to a sub-series level, where local information is more effectively captured, and computational costs are significantly reduced. Furthermore, a recent study \cite{dong2023simmtm} achieves superior fine-tuning performance by introducing a novel masked modeling task, involving the reconstruction of the original time series from multiple randomly masked series. Besides, TimeMAE \cite{cheng2023timemae} significantly surpasses previous competitive baselines in classification tasks by leveraging decoupled masked autoencoders to learn robust representations through two pretext tasks: masked codeword classification and masked representation regression. 

On the other hand, the forecast-as-pretraining schema has also been a subject of recent research. ForecastPFN \cite{dooley2023forecastpfn} introduces a prior-data fitted network trained on synthetic data and achieves accurate zero-shot forecasting on univariate time series. Besides, both TimeGPT-1 \cite{garza2023timegpt} and PreDcT \cite{timegpt2} explore the potential of training a foundation model for forecasting, yielding zero-shot forecasting capability under relatively short horizon lengths. Moreover, there is another related line of work focusing on adapting traditional pretrained generative language models to the time series domain, either through prompting \cite{gruver2023large} or fine-tuning \cite{fpt}. These approaches have demonstrated competitive results when compared to the traditional supervised approaches. \\

% , extracting general knowledge during the reconstruction process.

\begin{figure*}[htbp]
  \centering
  \includegraphics[width=\linewidth]{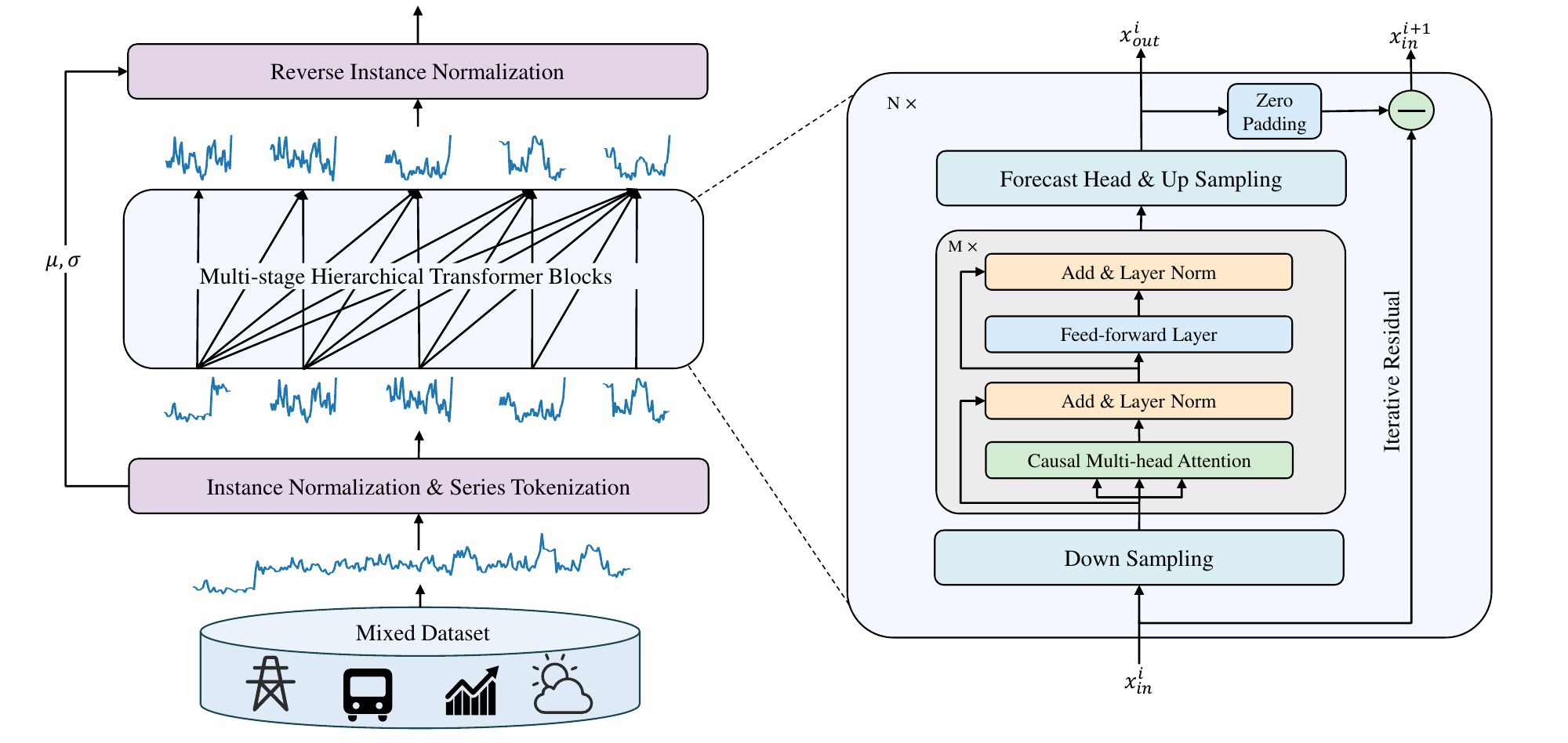}
  \caption{Illustration of the proposed GPHT model with two key features: (a) GPHT forecasts in an auto-regressive manner on the time series tokens. (b) Pretrained on the mixed dataset with the multi-stage hierarchical transformer blocks, GPHT excels in capturing the commonalities among time series originating from various data scenarios.}
  \label{fig:model}
\end{figure*}

Despite the effectiveness, existing methods exhibit two key shortcomings. Firstly, these models are typically trained on a single dataset with limited scale and patterns, impeding their ability to generalize to diverse forecasting scenarios. Secondly, both supervised and self-supervised approaches often necessitate a customized forecasting head, incurring multiple training costs under different horizon length settings. Regarding our proposed method, GPHT, it stands out as a generative self-supervised pretraining approach, distinguishing itself from existing methods by effectively addressing these issues. In detail, we investigate the feasibility of constructing a mixed dataset for training our model in an auto-regressive manner. This approach ensures the superior generalizability of GPHT, allowing it to be seamlessly adapted to any dataset, including unseen datasets, and forecast at arbitrary horizon lengths. Notably, experiments demonstrate that our method surpasses the baseline models across various fine-tuning and zero/few-shot learning settings in the traditional long-term forecasting task.

% Regarding our proposed method, GPHT, it stands out as a generative self-supervised pretraining approach, distinguishing itself from existing methods in two key aspects.

\section{Proposed Method}
In this section, we will delve into the specifics of the proposed GPHT method illustrated in Figure \ref{fig:model}, demonstrating its capacity to efficiently acquire precise time series forecasting through pretraining on the mixed dataset.

\subsection{Problem Definition}\label{sec:pd}
Given an input series $\mathcal{X}\in \mathbb{R}^{C\times L}$, a time series forecasting model is tasked with precisely predicting future values $\mathcal{Y}\in \mathbb{R}^{C\times H}$. Here, $L$, $H$, and $C$ represent the lookback window length, horizon length, and the number of channels, respectively.

GPHT adopts the channel-independent assumption \cite{DLinear, patchtst}, treating each multivariate time series as multiple independent univariate time series. In essence, GPHT conducts individual forecasting on each variate within the input series, and the resultant forecasts are concatenated to generate the final predictions. 

Moreover, we extend this methodology to the construction of the mixed pretraining dataset, where the heterogeneity of each variable is discarded and no extra information is taken into account. It can be therefore seamlessly applied to more diverse scenarios where the covariate information may be missing and the data itself may be synthetic. In practice, we concatenate the training segments from various real-world datasets to constitute the training set of the mixed dataset. This process is similarly applied to the validation and testing portions. Our approach ensures that GPHT is pretrained on a rich variety of temporal patterns, thereby enhancing its adaptability and generalization capabilities across diverse time series domains.

\subsection{Series Tokenization}\label{sec:preprocess}
Given that the majority of time series originate from signals captured by real-world sensors, the data inherently carry noise, and information is often sparsely distributed across the time points. Consequently, employing auto-regressive training on point-wise time series might yield suboptimal performance due to the risk of overfitting outliers and the associated error accumulation \cite{informer}, in addition to incurring high computation costs.

To address these challenges, we employ the series tokenization technique, a proven effective approach in time series modeling \cite{patchtst, crossformer, he2023shapewordnet}. Specifically, we adopt a non-overlapping tokenization strategy, reshaping the input series $\mathcal{X}$ into a sequence of time series tokens, $x \in \mathbb{R}^{C\times L'\times T}$, where $T \times L' = L$, $T$ represents the token length, and $L'$ can be considered as the sequence length. The tokenization strategy not only helps mitigate the impact of noise and sparse information distribution but also enhances the model's ability to better capture the local semantics, ultimately contributing to more robust and accurate time series forecasting.

Furthermore, recent research has highlighted a prevalent issue in time series data, characterized by a distribution shift \cite{DAIN, deng2021st, san} which means that the mean and variance of time series changes over time. This challenge significantly hinders the generalizability of deep-learning-based forecasters. The impact of this phenomenon is even more pronounced in the context of pretraining on a mixed dataset, where series inherently stem from distinct temporal distributions. To mitigate this issue, we introduce an Instance Normalization layer \cite{RevIN}, designed to address the distribution shift problem by normalizing the input series using the formula:
\begin{equation}
    x_{in} = \frac{x-\mu}{\sigma+\epsilon}\quad \text{and}\quad \mu = \mathbb{E}[\mathcal{X}], \sigma^2=\text{Var}[\mathcal{X}].
\end{equation}
Here, $\epsilon$ is a small constant, and $\mu, \sigma$ represent the instance-specific mean and standard deviation of $\mathcal{X}$, respectively. Subsequently, $x_{in}^0$ is fed into the model for further feature extraction. This normalization step enhances the model's robustness to distribution shifts, promoting more effective learning and improved generalization across diverse temporal patterns.

% Subsequently, $x$ is inputted into the model for further feature extraction.
% \subsection{Hierarchical Residual Learning}
% As depicted in Figure 1, the core of GPHT consists of multi-stage \textit{hierarchical transformer blocks} with a specially designed \textit{iterative residual learning} paradigm. 

\subsection{Hierarchical Transformer Blocks}
Multi-scale representation learning has demonstrated its effectiveness in various time series modeling tasks \cite{formertime, shabani2022scaleformer}, given the multiple periodic characteristics commonly found in real-world time series data. Furthermore, to better discover commonalities hidden within mixed datasets comprising various data scenarios, we posit the indispensability of a hierarchical encoder.

In practice, drawing inspiration from the multi-rate sampling strategy \cite{challu2022nhits}, we introduce a token-level multi-stage representation learning approach within our hierarchical transformer blocks: Suppose $x^i$ is the input series after tokenization of stage $i$, a max-pooling operation with a kernel size of $k_i$ is applied on each token, down-sampling the original data into $x_{in}^i \in \mathbb{R}^{C\times L' \times \frac{T}{k_i}}$. This operation retains a coarse-grained portion of the original data, compelling the encoder network to focus on modeling coarse patterns. Afterward, a standard multi-layer transformer network is employed for representation learning \cite{vaswani2017attention}:
\begin{equation}
    \begin{aligned}
        h_i^{in} &= PE_i(x_i^{in}) + Emb_i(x_i^{in}) \\
        h_i^{out} &= TRM_i(h_i^{in}).
    \end{aligned}
\end{equation}
Here, $PE$ represents a regular position embedding layer, and $Emb$ is a token projection layer projecting a time series token into the hidden space of the transformer. Besides, we employ the decoder-only transformers as the backbone, which incorporates causal attention masks to prevent information leakage during the auto-regressive generating process.

Finally, the learned hidden states $h_i^{out}$ are fed into the forecasting head, which is a linear layer, to predict future values for each token. In this regard, we adopt an up-sampling operation, which can be either linear interpolation or MLP, to map the predictions back to the original time series tokens, yielding the predictions $x_{out}^i$ of stage $i$. This process is represented as:
\begin{equation}
\label{eq:forecast}
    x_{out}^i = UpSampling(Forecast(h_i^{out})).
\end{equation}

This comprehensive approach to multi-stage representation learning and forecasting allows for the capturing of intricate temporal patterns at different scales, contributing to the model's effectiveness in handling diverse time series scenarios.

\subsection{Iterative Residual Learning}
Utilizing the multi-stage hierarchical transformer blocks, GPHT is expected to learn coarse-to-fine representations effectively. To fully leverage these representations, we propose a novel iterative residual learning strategy \cite{residual,oreshkin2019n}, transforming the forecasting process into an iterative approach.

Specifically, as illustrated in Figure 1, the input of stage $i+1$ is the residual of stage $i$'s input and output, defined as:
\begin{equation}
\label{res}
    x_{in}^{i+1} = x_{in}^i - PadFirstToken(x_{out}^i).
\end{equation}

Taking advantage of the auto-regressive training schema, each token in $x_{out}^i$ can be considered as the predicted value of the next token corresponding to the token in $x_{in}^i$. For the regression of the first token, we adopt a zero-padding for simplicity. Therefore, Equation \ref{res} effectively \textit{\textbf{refines the input for the next stage, eliminating the redundant information in the series}}. 

Intuitively, the pooling operation within the hierarchical transformers allows the model to concentrate on specific patterns at lower frequencies. Additionally, the task of deeper blocks is simplified as the shallower layers filter out the well-approximated signals. Therefore, the iterative residual learning strategy enables the model to focus on refining the finer details, allowing for the progressive enhancement of the predictive accuracy. Moreover, the strategy naturally suits the diverse patterns in the mixed pretraining dataset with adaptability to various temporal structures, thus guaranteeing promising generalizability.

\subsection{Optimization Target}
Let $S$ represent the number of stages, the intermediate forecasting result of GPHT is the sum of all the outputs of each hierarchical transformer block, defined as:
\begin{equation}
    y_{pred} = \sum_{i=1}^S x_{out}^i.
\end{equation}

Besides, a Reversed Instance Normalization layer is applied to the intermediate result through de-normalization, restoring the characteristics of the input series for better accuracy \cite{RevIN}:
\begin{equation}
    \mathcal{Y}_{pred} = y_{pred} \cdot (\sigma + \epsilon) + \mu.
\end{equation}

To fully leverage the mixed dataset and better capture temporal dependencies, we formulate the pretraining task as a standard language modeling task, employing a token-wise auto-regressive loss function as the optimization target (which is also a forecasting task). Using the same notations as in Section \ref{sec:pd}, and letting $H=T$, the optimization target of GPHT is:
\begin{equation}
    L=MSE(\mathcal{Y}_{pred}, Concat(\mathcal{X}[:,T:], \mathcal{Y}))
\end{equation}
Here, we use the standard Mean Squared Error (MSE) as the next-token-prediction loss function, as time series tokens are still continuous numerical values.

\subsection{Inference}\label{sec:inference}
Given that the pretraining task can be considered a forecasting task, \textit{\textbf{the pre-trained GPHT can be directly applied to downstream forecasting tasks without any modification}}. This sets our approach apart from mainstream pretraining methods \cite{yue2022ts2vec, cheng2023timemae, dong2023simmtm, fpt}, where a fine-tuning procedure is typically required.

On the other hand, we argue that the performance of GPHT can also be further enhanced through fine-tuning. In practice, to strike a balance between maintaining generalizability and improving performance on a specific dataset, we adopt a parameter-efficient tuning strategy. Specifically, only the forecasting heads described in Equation \ref{eq:forecast} are updated during the fine-tuning process. Importantly, these forecasting head parameters account for less than 0.5\% of the entire model.
% Since the pretraining task can be also regarded as a forecasting task, \textit{\textbf{the pre-trained GPHT can be directly applied in downstream forecasting tasks without any modification}}, which distinguishes our approach from mainstream pretraining methods \cite{yue2022ts2vec, cheng2023timemae, dong2023simmtm, fpt}, where a fine-tuning procedure is required. But also, the performance of GPHT can be boosted by fine-tuning. In practice, to both maintain the generalizability and improve the performance on a given dataset, we turn to a parameter-efficient tuning strategy, where only the forecasting heads described in Equation \ref{eq:forecast} are updated during the fine-tuning process, whose parameters account for less than 0.5\% of the entire model.

During inference, benefiting from the aforementioned training schema and the channel-independent assumption, GPHT is theoretically capable of conducting universal forecasting on any input multi-variate time series, regardless of arbitrary horizon lengths. The forecasting process is akin to the decoding process of a language model. Given any input $\mathcal{X}$, our model can initially predict the next first token. This predicted token is then concatenated to the end of the input series to generate predictions for the second token. Note that a max input length $L_m$ exists in our model due to the positional embeddings and heavy computation cost when addressing long sequences. Consequently, only the most recent $L_m$ tokens are fed into the model for forecasting.

\section{Experiments}
In this section, we conduct sufficient experiments on 8 widely used datasets in comparison with mainstream self-supervised pretraining methods and supervised methods to illustrate the effectiveness of our proposed GPHT.
\subsection{Experimental Setup}
\subsubsection{Datasets} Table \ref{tab:dataset} provides detailed descriptions of the used datasets, covering various data scenarios and scales. Among them, both \textbf{ETT}\footnote{https://github.com/zhouhaoyi/ETDataset} and \textbf{Electricity}\footnote{https://archive.ics.uci.edu/ml/datasets/ElectricityLoadDiagrams20112014} mainly records the consumption on electricity, and ETT can be further divided into 4 subsets according to the frequency. Besides, \textbf{Exchange}\footnote{https://github.com/laiguokun/multivariate-time-series-data} collects the daily exchange rate among 8 countries, \textbf{Traffic}\footnote{http://pems.dot.ca.gov} contains the data of traffic load sensors, and \textbf{Weather}\footnote{https://www.bgc-jena.mpg.de/wetter/} is made up of 21 climate indicators like air temperature. Following the standard protocol, we split each dataset into training, validation and testing sets according to the chronological order. The split ratio is 6:2:2 for the ETT dataset and 7:1:2 for the other datasets \cite{wu2021autoformer}.
\begin{table}[htbp]
  \centering
  % \small 
  \caption{The Statistics of Each Dataset.}
  \tabcolsep=0.1cm
  \label{tab:dataset}
    \begin{tabular}{llllll}
    \toprule
    \multicolumn{1}{c}{Dataset} & \multicolumn{1}{c}{Variables} & \multicolumn{1}{c}{Frequency} & \multicolumn{1}{c}{Length} & \multicolumn{1}{c}{Scope} \\
    \midrule
    \multicolumn{1}{c}{ETTh1/ETTh2} & \multicolumn{1}{c}{7} & \multicolumn{1}{c}{1 Hour} & \multicolumn{1}{c}{17420} & \multicolumn{1}{c}{Energy}\\
    \multicolumn{1}{c}{ETTm1/ETTm2} & \multicolumn{1}{c}{7} & \multicolumn{1}{c}{15 Minutes} & \multicolumn{1}{c}{69680} & \multicolumn{1}{c}{Energy}\\
    \multicolumn{1}{c}{Electricity} & \multicolumn{1}{c}{321} & \multicolumn{1}{c}{1 Hour} & \multicolumn{1}{c}{26304} & \multicolumn{1}{c}{Energy}\\
    \multicolumn{1}{c}{Exchange} & \multicolumn{1}{c}{8} & \multicolumn{1}{c}{1 Day} & \multicolumn{1}{c}{7588} & \multicolumn{1}{c}{Finance}\\
    \multicolumn{1}{c}{Traffic} & \multicolumn{1}{c}{862} & \multicolumn{1}{c}{1 Hour} & \multicolumn{1}{c}{17544} & \multicolumn{1}{c}{Transportation}\\
    \multicolumn{1}{c}{Weather} & \multicolumn{1}{c}{21} & \multicolumn{1}{c}{10 Minutes} & \multicolumn{1}{c}{52696} & \multicolumn{1}{c}{Weather}\\
    \midrule
    \end{tabular}%
\end{table}%
\subsubsection{Compared Baselines}
We select various state-of-the-art models as the baseline models in the experiments, encompassing both self-supervised and supervised approaches.

% Table generated by Excel2LaTeX from sheet 'Sheet2'
\begin{table*}[htbp]
  \tabcolsep=0.07cm
  \centering
  \caption{Multivariate time series forecasting results comparing GPHT with both SOTA self-supervised approaches and supervised approaches. The best results are in \textbf{bold} and the second best are \underline{underlined}.}\scalebox{0.9}{
    \begin{tabular}{c|c|cccc|cccccccc|cccccccc}
    \toprule
    \multicolumn{2}{c|}{Type} & \multicolumn{4}{c|}{Ours}     & \multicolumn{8}{c|}{Self-supervised}                          & \multicolumn{8}{c}{Supervised} \\
    \multicolumn{2}{c|}{Methods} & \multicolumn{2}{c}{GPHT*} & \multicolumn{2}{c|}{GPHT} & \multicolumn{2}{c}{PatchTST} & \multicolumn{2}{c}{FPT} & \multicolumn{2}{c}{SimMTM} & \multicolumn{2}{c|}{TimeMAE} & \multicolumn{2}{c}{PatchTST} & \multicolumn{2}{c}{iTransformer} & \multicolumn{2}{c}{TimesNet} & \multicolumn{2}{c}{DLinear} \\
    \multicolumn{2}{c|}{Metric} & MSE   & MAE   & MSE   & MAE   & MSE   & MAE   & MSE   & MAE   & MSE   & MAE   & MSE   & MAE   & MSE   & MAE   & MSE   & MAE   & MSE   & MAE   & MSE   & MAE \\
    \midrule
    \multirow{4}[2]{*}{\rotatebox{90}{Electricity}} & 96    & \textbf{0.128 } & \textbf{0.219 } & \textbf{0.128 } & \textbf{0.219 } & \underline{0.132}  & 0.225  & 0.139  & 0.238  & 0.133  & \underline{0.223}  & 0.133  & 0.230  & 0.138  & 0.233  & \underline{0.132}  & 0.228  & 0.177  & 0.281  & 0.141  & 0.238  \\
          & 192   & \underline{0.147}  & \textbf{0.236 } & \textbf{0.146 } & \textbf{0.236 } & 0.148  & 0.241  & 0.155  & 0.252  & \underline{0.147}  & \underline{0.237}  & 0.150  & 0.246  & 0.153  & 0.247  & 0.154  & 0.249  & 0.193  & 0.295  & 0.154  & 0.251  \\
          & 336   & \textbf{0.165 } & \textbf{0.255 } & \textbf{0.165 } & \textbf{0.255 } & 0.167  & \underline{0.260}  & 0.170  & 0.267  & \underline{0.166}  & 0.265  & \underline{0.166}  & 0.265  & 0.170  & 0.263  & 0.172  & 0.267  & 0.206  & 0.306  & 0.170  & 0.269  \\
          & 720   & 0.206  & \textbf{0.292 } & 0.207  & \textbf{0.292 } & 0.205  & \textbf{0.292 } & 0.208  & 0.299  & \underline{0.203}  & 0.297  & \textbf{0.199 } & 0.296  & 0.206  & \underline{0.295}  & 0.204  & 0.296  & 0.223  & 0.320  & 0.205  & 0.302  \\
    \midrule
    \multirow{4}[2]{*}{\rotatebox{90}{Exchange}} & 96    & 0.096  & \underline{0.216}  & \textbf{0.087 } & \textbf{0.207 } & \underline{0.088}  & \textbf{0.207 } & 0.098  & 0.222  & 0.100  & 0.226  & 0.229  & 0.352  & 0.094  & \underline{0.216}  & 0.099  & 0.225  & 0.166  & 0.305  & \textbf{0.087 } & 0.217  \\
          & 192   & 0.183  & 0.304  & \underline{0.172}  & \textbf{0.296 } & 0.186  & 0.308  & 0.209  & 0.327  & 0.210  & 0.332  & 0.653  & 0.581  & 0.191  & 0.311  & 0.206  & 0.329  & 0.303  & 0.413  & \textbf{0.164 } & \underline{0.298}  \\
          & 336   & \underline{0.322}  & \underline{0.410}  & \textbf{0.309 } & \textbf{0.400 } & 0.374  & 0.446  & 0.398  & 0.463  & 0.389  & 0.460  & 1.524  & 0.887  & 0.343  & 0.427  & 0.370  & 0.448  & 0.445  & 0.511  & 0.333  & 0.437  \\
          & 720   & \underline{0.833}  & \underline{0.685}  & \textbf{0.808 } & \textbf{0.669 } & 0.857  & 0.692  & 1.010  & 0.747  & 1.104  & 0.800  & 2.525  & 1.193  & 0.888  & 0.706  & 0.963  & 0.746  & 1.389  & 0.899  & 0.988  & 0.749  \\
    \midrule
    \multirow{4}[2]{*}{\rotatebox{90}{Traffic}} & 96    & \underline{0.348}  & \underline{0.236}  & \textbf{0.346 } & \textbf{0.234 } & 0.382  & 0.262  & 0.388  & 0.279  & 0.368  & 0.262  & 0.365  & 0.252  & 0.395  & 0.272  & 0.361  & 0.266  & 0.600  & 0.323  & 0.411  & 0.284  \\
          & 192   & 0.374  & \underline{0.248}  & \textbf{0.371 } & \textbf{0.246 } & 0.385  & 0.261  & 0.411  & 0.287  & \underline{0.373}  & 0.251  & 0.383  & 0.260  & 0.411  & 0.278  & 0.378  & 0.271  & 0.612  & 0.327  & 0.423  & 0.289  \\
          & 336   & 0.392  & 0.259  & \textbf{0.388 } & \underline{0.256}  & 0.409  & 0.275  & 0.423  & 0.293  & 0.395  & \textbf{0.254 } & 0.399  & 0.269  & 0.424  & 0.284  & \underline{0.390}  & 0.274  & 0.628  & 0.344  & 0.437  & 0.297  \\
          & 720   & 0.428  & \underline{0.284}  & \textbf{0.423 } & \textbf{0.279 } & 0.438  & 0.291  & 0.449  & 0.307  & 0.432  & 0.290  & 0.438  & 0.291  & 0.453  & 0.300  & \underline{0.424}  & 0.291  & 0.657  & 0.349  & 0.467  & 0.316  \\
    \midrule
    \multirow{4}[2]{*}{\rotatebox{90}{Weather}} & 96    & 0.155  & \textbf{0.196 } & 0.154  & \textbf{0.196 } & \underline{0.148}  & \textbf{0.196 } & 0.152  & 0.201  & 0.152  & 0.201  & 0.151  & 0.208  & \textbf{0.147 } & \underline{0.197}  & 0.162  & 0.212  & 0.168  & 0.225  & 0.176  & 0.236  \\
          & 192   & 0.203  & \textbf{0.240 } & 0.201  & \textbf{0.240 } & \underline{0.193}  & \textbf{0.240 } & 0.197  & \underline{0.244}  & 0.198  & 0.245  & 0.198  & 0.256  & \textbf{0.191 } & \textbf{0.240 } & 0.205  & 0.251  & 0.218  & 0.268  & 0.217  & 0.275  \\
          & 336   & 0.259  & 0.283  & 0.257  & 0.283  & \textbf{0.244 } & \textbf{0.279 } & 0.252  & 0.287  & 0.249  & 0.285  & \underline{0.246}  & 0.294  & \textbf{0.244 } & \underline{0.282}  & 0.257  & 0.291  & 0.269  & 0.301  & 0.264  & 0.315  \\
          & 720   & 0.338  & 0.337  & 0.335  & 0.337  & 0.321  & \textbf{0.334 } & 0.329  & 0.340  & 0.324  & \underline{0.335}  & \textbf{0.316 } & 0.351  & \underline{0.320}  & \textbf{0.334 } & 0.325  & 0.337  & 0.340  & 0.350  & 0.325  & 0.364  \\
    \midrule
    \multirow{4}[2]{*}{\rotatebox{90}{ETTh1}} & 96    & 0.378  & \underline{0.388}  & \textbf{0.363 } & \textbf{0.382 } & 0.384  & 0.401  & 0.388  & 0.405  & 0.383  & 0.411  & 0.431  & 0.450  & 0.382  & 0.403  & 0.405  & 0.419  & 0.421  & 0.438  & \underline{0.375}  & 0.396  \\
          & 192   & 0.425  & \underline{0.416}  & \textbf{0.405 } & \textbf{0.408 } & 0.427  & 0.431  & 0.422  & 0.423  & 0.417  & 0.432  & 0.484  & 0.486  & \underline{0.416}  & 0.423  & 0.448  & 0.447  & 0.482  & 0.479  & 0.428  & 0.437  \\
          & 336   & 0.456  & \underline{0.432}  & \underline{0.430}  & \textbf{0.423 } & 0.461  & 0.450  & 0.442  & 0.435  & \textbf{0.425 } & 0.439  & 0.515  & 0.507  & 0.441  & 0.440  & 0.482  & 0.470  & 0.528  & 0.505  & 0.448  & 0.449  \\
          & 720   & 0.454  & \underline{0.449}  & \textbf{0.414 } & \textbf{0.435 } & 0.460  & 0.465  & 0.469  & 0.473  & \underline{0.437}  & 0.456  & 0.595  & 0.577  & 0.470  & 0.475  & 0.560  & 0.537  & 0.527  & 0.510  & 0.505  & 0.514  \\
    \midrule
    \multirow{4}[2]{*}{\rotatebox{90}{ETTh2}} & 96    & 0.307  & 0.347  & 0.296  & \textbf{0.340 } & 0.297  & 0.354  & \underline{0.291}  & 0.349  & 0.298  & 0.350  & 0.294  & 0.358  & \textbf{0.286 } & \underline{0.342}  & 0.305  & 0.361  & 0.355  & 0.408  & 0.296  & 0.360  \\
          & 192   & 0.373  & 0.389  & 0.363  & \textbf{0.384 } & 0.388  & 0.406  & \underline{0.356}  & 0.390  & 0.360  & \underline{0.388}  & \textbf{0.352 } & 0.397  & 0.357  & 0.389  & 0.391  & 0.412  & 0.403  & 0.434  & 0.391  & 0.423  \\
          & 336   & 0.399  & 0.414  & 0.392  & \underline{0.410}  & 0.392  & 0.413  & \underline{0.387}  & 0.418  & 0.388  & \underline{0.410}  & 0.394  & 0.427  & \textbf{0.377 } & \textbf{0.409 } & 0.418  & 0.433  & 0.398  & 0.434  & 0.445  & 0.460  \\
          & 720   & 0.412  & \underline{0.429}  & \underline{0.407}  & \textbf{0.427 } & 0.413  & 0.442  & 0.415  & 0.448  & 0.412  & 0.435  & 0.539  & 0.510  & \textbf{0.406 } & 0.440  & 0.437  & 0.455  & 0.443  & 0.465  & 0.700  & 0.592  \\
    \midrule
    \multirow{4}[2]{*}{\rotatebox{90}{ETTm1}} & 96    & 0.301  & 0.345  & 0.291  & \textbf{0.339 } & \textbf{0.281 } & \underline{0.341}  & \underline{0.290}  & 0.346  & 0.296  & 0.349  & 0.301  & 0.348  & 0.298  & 0.345  & 0.306  & 0.360  & 0.331  & 0.372  & 0.303  & 0.346  \\
          & 192   & 0.347  & 0.374  & 0.337  & \textbf{0.368 } & \textbf{0.326 } & 0.372  & \underline{0.330}  & \underline{0.371}  & 0.334  & 0.373  & 0.351  & 0.383  & 0.339  & 0.374  & 0.345  & 0.382  & 0.435  & 0.421  & 0.338  & \textbf{0.368 } \\
          & 336   & 0.388  & 0.401  & 0.377  & \underline{0.393}  & \textbf{0.348 } & \textbf{0.384 } & \underline{0.366}  & \underline{0.393}  & 0.371  & 0.398  & 0.390  & 0.408  & 0.381  & 0.401  & 0.378  & 0.402  & 0.457· & 0.445  & 0.373  & \underline{0.393 } \\
          & 720   & 0.465  & 0.441  & 0.452  & 0.433  & \textbf{0.399 } & \textbf{0.418 } & \underline{0.416}  & \underline{0.421}  & 0.418  & 0.425  & 0.457  & 0.446  & 0.428  & 0.431  & 0.443  & 0.439  & 0.526  & 0.481  & 0.428  & 0.423  \\
    \midrule
    \multirow{4}[2]{*}{\rotatebox{90}{ETTm2}} & 96    & 0.179  & \underline{0.257}  & \textbf{0.170 } & \textbf{0.250 } & \underline{0.171}  & \underline{0.257}  & \underline{0.171}  & 0.261  & 0.173  & 0.264  & 0.180  & 0.267  & 0.174  & 0.261  & 0.174  & 0.266  & 0.190  & 0.276  & \textbf{0.170 } & 0.264  \\
          & 192   & 0.242  & \underline{0.298}  & \textbf{0.230 } & \textbf{0.291 } & 0.236  & 0.304  & \underline{0.231}  & 0.302  & \textbf{0.230 } & 0.299  & 0.243  & 0.312  & 0.238  & 0.307  & 0.247  & 0.315  & 0.244  & 0.311  & 0.233  & 0.311  \\
          & 336   & 0.300  & 0.334  & \underline{0.285}  & \textbf{0.327 } & 0.291  & 0.344  & 0.288  & 0.343  & \textbf{0.282 } & \underline{0.332}  & 0.308  & 0.355  & 0.293  & 0.346  & 0.292  & 0.343  & 0.302  & 0.349  & 0.298  & 0.358  \\
          & 720   & 0.400  & 0.393  & 0.380  & \textbf{0.386 } & 0.388  & 0.404  & 0.389  & 0.406  & \underline{0.374}  & \underline{0.390}  & 0.395  & 0.407  & \textbf{0.373 } & 0.401  & 0.375  & 0.395  & 0.406  & 0.406  & 0.423  & 0.437  \\
    \midrule
    \multicolumn{2}{c|}{\#1 Counts} & \multicolumn{2}{c}{8} & \multicolumn{2}{c|}{41} & \multicolumn{2}{c}{13} & \multicolumn{2}{c}{0} & \multicolumn{2}{c}{4} & \multicolumn{2}{c|}{3} & \multicolumn{2}{c}{10} & \multicolumn{2}{c}{0} & \multicolumn{2}{c}{0} & \multicolumn{2}{c}{4} \\
    \bottomrule
    \end{tabular}}%
  \label{tab:main_results}%
\end{table*}%

Among them, \textbf{FPT \cite{fpt}} introduces a framework that utilizes parameter-efficient fine-tuning on pretrained generative language models, adapting them to time series tasks. \textbf{SimMTM \cite{dong2023simmtm}} reframes the standard masked time series modeling target into recovering masked time points through the weighted aggregation of multiple neighbors. Additionally, \textbf{TimeMAE \cite{cheng2023timemae}} leverages decoupled masked autoencoders to learn robust representations through masked codeword classification and masked representation regression. On the other hand, we include superior supervised models to better demonstrate the effectiveness of GPHT, including Transformer-based models such as \textbf{PatchTST \cite{patchtst}} and \textbf{iTransformer \cite{liu2023itransformer}}, a linear model-based approach \textbf{DLinear \cite{DLinear}}, and a convolution-based model \textbf{TimesNet \cite{wu2022timesnet}}, which is also a multi-scale model. The performance of these methods effectively represents the utmost accuracy achievable by current forecasting models.
\subsubsection{Implementation Details}
We employ ADAM as the default optimizer throughout all the experiments and use mean squared error (MSE) and mean absolute error (MAE) as the evaluation metrics. A lower MSE/MAE indicates better performance. For the baseline models, we implement them with official codes and recommended parameter settings.

As for GPHT, we set the token length $T$ to 48, and the max input length $L_m$ is set to 7 to accommodate the lookback window length. The model comprises 4 stages of hierarchical transformer blocks, each with three-layer decoder-only transformers. The down-sampling ratio for each stage is set to $[8,4,2,1]$ respectively. All experiments are conducted for three runs with a fixed random seed on a single NVIDIA RTX 4090 24GB GPU.

%   \label{tab:main_results}%

\subsection{Main Results}
We report the long-term multi-variate forecasting results in Table \ref{tab:main_results}. For a fair comparison, the lookback window length $L$ is set to 336 for every model and dataset, and the horizon length $H$ is $[96,192,336,720]$ following the standard protocol. Here, GPHT* denotes our model pretrained on the mixed dataset without any modification, while GPHT represents the fine-tuned version for each dataset, as described in Section \ref{sec:inference}.

As shown in the table, we can draw some interesting conclusions. Firstly, self-supervised methods showcase competitive performance with their supervised counterparts, underscoring the efficacy of advanced pretraining techniques. Secondly, the tokenization technique emerges as a crucial factor in achieving precise forecasting, as evidenced by the outstanding performance of both FPT and patchTST among the baseline models. Finally, it appears that modern forecasters have approached the precision limits of some benchmark datasets, likely due to inherent noise and unpredictable distribution shifts in the data. This phenomenon results in minimal differences in outcomes among various models.

On the other hand, in terms of our proposed GPHT, it consistently surpasses its competitors across most experimental settings. The tuned GPHT model establishes the most accurate forecasting under over \textbf{65\%} cases with its powerful counterparts. Specifically, GPHT exhibits an average MSE reduction of \textbf{9.23\%} on the Exchange dataset, \textbf{1.60\%} on the Traffic dataset, and \textbf{3.00\%} on the ETTh1 dataset, in comparison with the best baseline under all experimental forecasting lengths. Regarding MAE evaluation, the improvements are more pronounced at \textbf{5.30\%, 3.97\%}, and \textbf{5.07\%} respectively. Although the relative improvements under certain settings may not be as substantial, as we claimed before, the consistently superior performance provides strong evidence for the effectiveness of our approach. This further demonstrates that our model can better capture temporal dependencies in various data scenarios, benefiting from its pretraining on the mixed dataset.

Besides, it is noteworthy that our model exhibits superior performance at relatively shorter horizon lengths. Specifically, GPHT surpasses PatchTST on every dataset when $H=96$, resulting in an average MAE reduction of \textbf{4.55\%}. We attribute this performance boost to the explicit modeling of temporal dependencies in the output series.  On the other hand, due to the inevitable error accumulation issue caused by the auto-regressive forecasting schema, the superiority of GPHT decreases with longer $H$s and the the average reduction of MAE comes to \textbf{3.38\%} when $H=720$.

Even more surprisingly, GPHT*, representing the model after pretraining \textbf{without any fine-tuning or modification}, proves to be competitive with the baseline models. Specifically, GPHT* achieves the best or the second-best performance in \textbf{26} out of all 64 settings. When compared to a single model, it outperforms FPT and supervised PatchTST under \textbf{44} and \textbf{40} experimental settings, respectively. This result validates the feasibility of learning the commonalities of different time series by training on the mixed dataset with the channel-independent assumption. It also underscores the incredible generalizability of our proposed model, primarily owing to the specially designed multi-stage hierarchical blocks and the iterative residual learning strategy.

\subsection{Zero-shot Evaluation}
To further highlight GPHT's capacity to learn general knowledge and discover common patterns from the mixed dataset, we conduct zero-shot forecasting experiments on the Exchange, Weather, and Traffic datasets, originating from various data scenarios with distinct scales. The zero-shot forecasting task is conceptually challenging for methods that model cross-variable dependencies, hence, only the channel-independent models are considered as baseline models. It is important to note that ForecastPFN \cite{dooley2023forecastpfn} is specifically designed for zero-shot forecasting, but its performance is heavily contingent on how the training data is synthesized, and as such, it is not included in this experiment. The models are trained on the mixed dataset comprised of the remaining 7 datasets and evaluated directly on the target dataset. The results are presented in Table \ref{tab:zero-shot}.

% Table generated by Excel2LaTeX from sheet 'Sheet2'
\begin{table}[htbp]
  \centering
  \tabcolsep=0.07cm
  \caption{Comparison on zero-shot forecasting task. The best results are highlighted in \textbf{bold}. }
    \begin{tabular}{c|c|cc|cc|cc|cc}
    \toprule
    \multicolumn{2}{c|}{Methods} & \multicolumn{2}{c|}{GPHT} & \multicolumn{2}{c|}{FPT} & \multicolumn{2}{c|}{PatchTST} & \multicolumn{2}{c}{DLinear} \\
    \multicolumn{2}{c|}{Metric} & MSE & MAE & MSE & MAE & MSE & MAE & MSE & MAE \\
    \midrule
    \multirow{4}[2]{*}{\rotatebox{90}{Exchange}} & 96 & \textbf{0.098 } & \textbf{0.219 } & 0.104  & 0.226  & 0.102  & 0.227  & 0.169  & 0.316  \\
       & 192 & \textbf{0.183 } & \textbf{0.305 } & 0.218  & 0.333  & 0.205  & 0.325  & 0.230  & 0.374  \\
       & 336 & \textbf{0.321 } & \textbf{0.411 } & 0.391  & 0.460  & 0.362  & 0.440  & 0.334  & 0.444  \\
       & 720 & 0.824  & 0.682  & 0.978  & 0.734  & 0.991  & 0.745  & \textbf{0.560 } & \textbf{0.591 } \\
    \midrule
    \multirow{4}[2]{*}{\rotatebox{90}{Traffic}} & 96 & \textbf{0.411 } & \textbf{0.291 } & 0.447  & 0.331  & 0.433  & 0.314  & 0.453  & 0.328  \\
       & 192 & \textbf{0.435 } & \textbf{0.302 } & 0.461  & 0.335  & 0.447  & 0.319  & 0.464  & 0.330  \\
       & 336 & \textbf{0.460 } & \textbf{0.316 } & 0.477  & 0.343  & 0.465  & 0.329  & 0.481  & 0.340  \\
       & 720 & 0.521  & 0.353  & \textbf{0.503 } & 0.356  & 0.504  & 0.354  & 0.506  & \textbf{0.351 } \\
    \midrule
    \multirow{4}[2]{*}{\rotatebox{90}{Weather}} & 96 & \textbf{0.202 } & \textbf{0.244 } & 0.216  & 0.264  & 0.207  & 0.259  & 0.239  & 0.297  \\
       & 192 & \textbf{0.248 } & \textbf{0.283 } & 0.260  & 0.301  & 0.257  & 0.299  & 0.275  & 0.325  \\
       & 336 & \textbf{0.306 } & \textbf{0.324 } & 0.328  & 0.351  & 0.340  & 0.350  & 0.323  & 0.360  \\
       & 720 & \textbf{0.389 } & \textbf{0.377 } & 0.414  & 0.403  & 0.414  & 0.402  & 0.392  & 0.405  \\
    \bottomrule
    \end{tabular}%
  \label{tab:zero-shot}%
\end{table}%

Clearly, GPHT consistently outperforms other models across various settings, showcasing pronounced relative improvements. We attribute this success primarily to the multi-stage hierarchical transformer blocks, designed to capture diverse temporal patterns with different resolutions. Consequently, GPHT exhibits better transferability to unseen time series. Additionally, all models achieve forecasting accuracy at an acceptable error level, with DLinear even demonstrating unprecedented precision on the Exchange dataset when $H=720$. These results strongly validate the feasibility of our approach to learning commonalities in time series through pretraining on a mixed dataset.

\subsection{Few-shot Evaluation}
% Table generated by Excel2LaTeX from sheet 'Sheet2'
\begin{table*}[htbp]
  \centering
  \small
  \tabcolsep=0.07cm
  \caption{Multivariate forecasting results under few-shot learning settings. The best results are highlighted in bold.}
    \begin{tabular}{c|c|cccccccccc|cccccccccc}
    \toprule
    \multicolumn{2}{c|}{Portion} & \multicolumn{10}{c|}{5\%}                       & \multicolumn{10}{c}{10\%} \\
\cmidrule{3-22}    \multicolumn{2}{c|}{Methods} & \multicolumn{2}{c}{GPHT} & \multicolumn{2}{c}{FPT} & \multicolumn{2}{c}{SimMTM} & \multicolumn{2}{c}{PatchTST} & \multicolumn{2}{c|}{iTransformer} & \multicolumn{2}{c}{GPHT} & \multicolumn{2}{c}{FPT} & \multicolumn{2}{c}{SimMTM} & \multicolumn{2}{c}{PatchTST} & \multicolumn{2}{c}{iTransformer} \\
    \multicolumn{2}{c|}{Metric} & MSE & MAE & MSE & MAE & MSE & MAE & MSE & MAE & MSE & MAE & MSE & MAE & MSE & MAE & MSE & MAE & MSE & MAE & MSE & MAE \\
    \midrule
    \multirow{4}[2]{*}{\rotatebox{90}{Electricity}} & 96 & \textbf{0.143 } & \textbf{0.237 } & 0.148  & 0.246  & 0.152  & 0.255  & 0.188  & 0.292  & 0.155  & 0.256  & \textbf{0.140 } & \textbf{0.233 } & 0.149  & 0.248  & 0.146  & 0.246  & 0.147  & 0.245  & 0.148  & 0.247  \\
       & 192 & \textbf{0.162 } & \textbf{0.254 } & 0.163  & 0.259  & 0.167  & 0.268  & 0.202  & 0.304  & 0.172  & 0.272  & \textbf{0.159 } & \textbf{0.250 } & 0.164  & 0.261  & 0.163  & 0.262  & 0.162  & 0.258  & 0.167  & 0.266  \\
       & 336 & 0.184  & \textbf{0.275 } & \textbf{0.181 } & 0.277  & 0.187  & 0.287  & 0.219  & 0.318  & 0.197  & 0.295  & \textbf{0.180 } & \textbf{0.271 } & 0.183  & 0.280  & 0.184  & 0.280  & 0.181  & 0.276  & 0.192  & 0.290  \\
       & 720 & 0.238  & 0.321  & \textbf{0.231 } & \textbf{0.315 } & 0.240  & 0.326  & 0.264  & 0.351  & 0.261  & 0.344  & 0.231  & \textbf{0.313 } & 0.234  & 0.318  & 0.242  & 0.325  & \textbf{0.230 } & 0.315  & 0.244  & 0.329  \\
    \midrule
    \multirow{4}[2]{*}{\rotatebox{90}{ETTh1}} & 96 & \textbf{0.383 } & \textbf{0.390 } & 0.478  & 0.474  & 0.537  & 0.502  & 0.505  & 0.481  & 0.580  & 0.520  & \textbf{0.382 } & \textbf{0.391 } & 0.453  & 0.454  & 0.482  & 0.467  & 0.450  & 0.448  & 0.557  & 0.514  \\
       & 192 & \textbf{0.426 } & \textbf{0.416 } & 0.705  & 0.577  & 0.580  & 0.525  & 0.576  & 0.514  & 0.670  & 0.557  & \textbf{0.424 } & \textbf{0.418 } & 0.522  & 0.494  & 0.532  & 0.498  & 0.523  & 0.489  & 0.668  & 0.562  \\
       & 336 & \textbf{0.453 } & \textbf{0.430 } & 0.736  & 0.571  & 0.603  & 0.543  & 0.672  & 0.554  & 0.726  & 0.577  & \textbf{0.450 } & \textbf{0.443 } & 0.571  & 0.522  & 0.561  & 0.523  & 0.523  & 0.494  & 0.684  & 0.559  \\
       & 720 & \textbf{0.433 } & \textbf{0.440 } & 0.718  & 0.579  & 0.708  & 0.597  & 0.759  & 0.625  & 0.802  & 0.626  & \textbf{0.427 } & \textbf{0.442 } & 0.574  & 0.535  & 0.734  & 0.617  & 0.508  & 0.502  & 0.709  & 0.587  \\
    \midrule
    \multirow{4}[2]{*}{\rotatebox{90}{ETTh2}} & 96 & \textbf{0.298 } & \textbf{0.343 } & 0.476  & 0.457  & 0.381  & 0.401  & 0.502  & 0.475  & 0.395  & 0.420  & \textbf{0.298 } & \textbf{0.343 } & 0.330  & 0.371  & 0.332  & 0.373  & 0.320  & 0.366  & 0.365  & 0.398  \\
       & 192 & \textbf{0.368 } & \textbf{0.386 } & 0.714  & 0.573  & 0.435  & 0.435  & 0.569  & 0.511  & 0.448  & 0.453  & \textbf{0.367 } & \textbf{0.387 } & 0.419  & 0.422  & 0.391  & 0.411  & 0.400  & 0.416  & 0.432  & 0.439  \\
       & 336 & \textbf{0.402 } & \textbf{0.412 } & 0.683  & 0.573  & 0.431  & 0.441  & 0.540  & 0.506  & 0.453  & 0.462  & \textbf{0.396 } & \textbf{0.413 } & 0.419  & 0.434  & 0.410  & 0.429  & 0.405  & 0.425  & 0.437  & 0.450  \\
       & 720 & \textbf{0.417 } & \textbf{0.429 } & 0.648  & 0.557  & 0.450  & 0.459  & 0.506  & 0.494  & 0.483  & 0.484  & \textbf{0.409 } & \textbf{0.428 } & 0.506  & 0.485  & 0.448  & 0.460  & 0.483  & 0.474  & 0.463  & 0.471  \\
    \midrule
    \multirow{4}[2]{*}{\rotatebox{90}{ETTm1}} & 96 & 0.513  & 0.438  & 0.395  & 0.409  & 0.446  & 0.434  & \textbf{0.376 } & \textbf{0.395 } & 0.434  & 0.436  & 0.506  & 0.427  & 0.403  & 0.411  & 0.442  & 0.430  & \textbf{0.386 } & \textbf{0.401 } & 0.420  & 0.427  \\
       & 192 & 0.552  & 0.464  & 0.410  & 0.417  & 0.461  & 0.436  & \textbf{0.391 } & \textbf{0.402 } & 0.472  & 0.456  & 0.563  & 0.458  & 0.430  & 0.426  & 0.454  & 0.431  & \textbf{0.406 } & \textbf{0.413 } & 0.472  & 0.456  \\
       & 336 & 0.609  & 0.491  & 0.453  & 0.440  & 0.500  & 0.455  & \textbf{0.438 } & \textbf{0.431 } & 0.531  & 0.486  & 0.634  & 0.492  & 0.472  & 0.443  & 0.552  & 0.469  & \textbf{0.438 } & \textbf{0.429 } & 0.530  & 0.486  \\
       & 720 & 0.685  & 0.581  & 0.742  & 0.566  & 0.590  & 0.503  & \textbf{0.549 } & \textbf{0.495 } & 0.615  & 0.527  & 0.721  & 0.534  & 0.665  & 0.526  & 0.715  & 0.539  & \textbf{0.499 } & \textbf{0.464 } & 0.629  & 0.533  \\
    \midrule
    \multirow{4}[2]{*}{\rotatebox{90}{ETTm2}} & 96 & \textbf{0.186 } & \textbf{0.271 } & 0.196  & 0.278  & 0.216  & 0.293  & 0.196  & 0.276  & 0.211  & 0.295  & \textbf{0.173 } & \textbf{0.256 } & 0.198  & 0.275  & 0.203  & 0.283  & 0.191  & 0.270  & 0.198  & 0.284  \\
       & 192 & \textbf{0.248 } & \textbf{0.311 } & 0.263  & 0.316  & 0.267  & 0.324  & 0.258  & 0.315  & 0.269  & 0.322  & \textbf{0.234 } & \textbf{0.297 } & 0.263  & 0.315  & 0.256  & 0.315  & 0.252  & 0.308  & 0.254  & 0.318  \\
       & 336 & \textbf{0.307 } & \textbf{0.347 } & 0.336  & 0.363  & 0.315  & 0.356  & 0.318  & 0.353  & 0.325  & 0.370  & \textbf{0.293 } & \textbf{0.335 } & 0.320  & 0.350  & 0.305  & 0.345  & 0.310  & 0.345  & 0.305  & 0.352  \\
       & 720 & 0.412  & 0.409  & 0.453  & 0.430  & \textbf{0.406 } & \textbf{0.406 } & 0.447  & 0.427  & 0.441  & 0.434  & 0.398  & \textbf{0.395 } & 0.426  & 0.412  & \textbf{0.397}  & 0.398  & 0.398  & 0.397  & 0.405  & 0.408  \\
    \bottomrule
    \end{tabular}%
  \label{tab:few-shot}%
\end{table*}%
In real-world applications, the initial observation of time series may be of a limited size, posing challenges for training accurate forecasters. To evaluate the representation power of GPHT under such circumstances, we conduct few-shot evaluations on the ETT and Electricity dataset. In detail, only a portion (10\% or 5\%, following existing work \cite{fpt}) of the training instances are used for training the models, and we evaluate their MSE and MAE on the full test set. The results are reported in Table \ref{tab:few-shot}.

In comparison to the selected baseline models, GPHT consistently achieves superior performance across various experimental settings, particularly on small-scale datasets. Specifically, when compared to the best-performing baseline in the 10\% setting, GPHT demonstrates a relative average MSE reduction of \textbf{16.02\%} and \textbf{7.02\%} on the ETTh1 and ETTh2 datasets, respectively. With a further reduction in the training data to only 5\%, the relative improvements become even more pronounced, reaching \textbf{30.19\%} and \textbf{12.49\%}. However, it is noteworthy that GPHT exhibits poor forecasting performance on the ETTm1 dataset. We believe this observation may be attributed to the potential heterogeneity of ETTm1 compared to other datasets, where the pretraining procedure might compromise generalizability when insufficient data is available. Addressing how to identify and leverage dataset heterogeneities for enhanced pretraining remains a subject for future exploration.

\subsection{Ablation Study}
\subsubsection{Hierarchical Architecture} 
\begin{figure}[htbp]
  \centering
  \includegraphics[width=\linewidth]{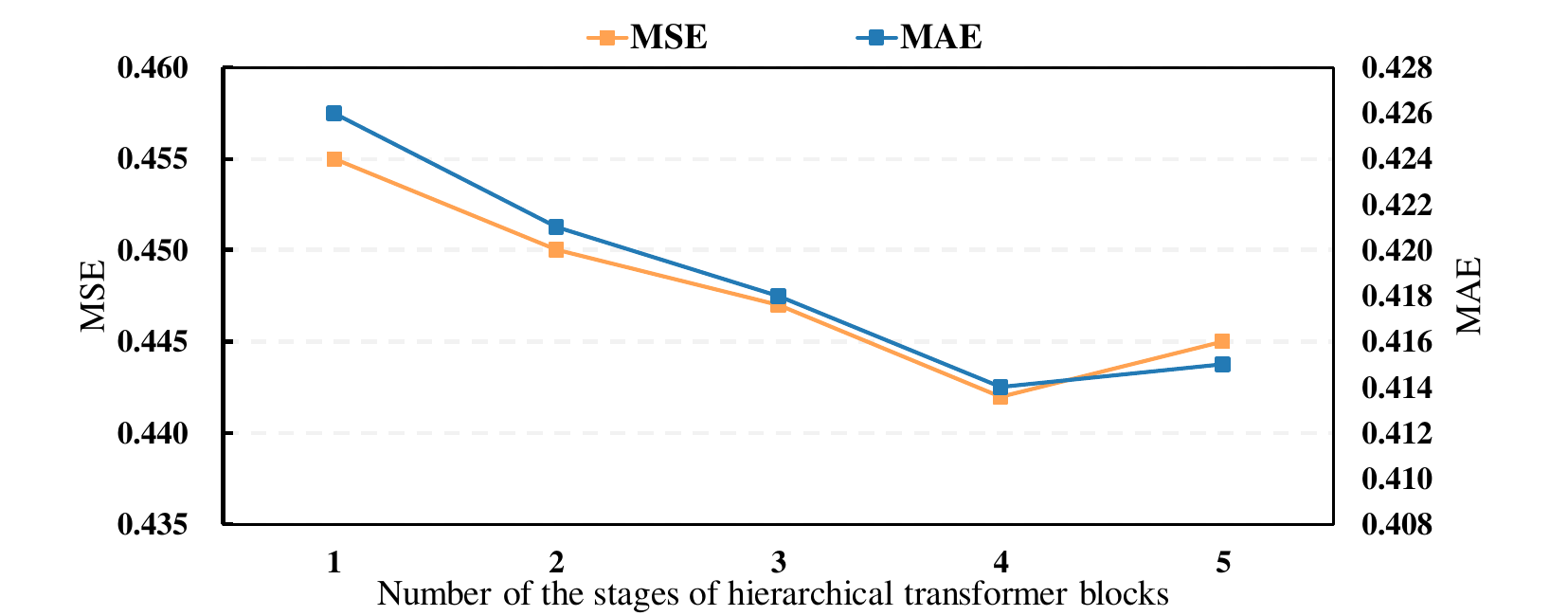}
  \caption{Performance comparison between GHPT with different stages of hierarchical transformer blocks.}
  \label{fig:stage}
\end{figure}

In this section, we explore the influence of hierarchical transformer blocks on GPHT's performance. We present the averaged MSE and MAE evaluations for GPHT with varying stages of hierarchical transformer blocks across all benchmark datasets, considering a forecasting horizon of $H=720$ (see Figure \ref{fig:stage}). As the number of stages increases, GPHT is theoretically better equipped to capture diverse temporal dependencies within the mixed dataset, such as different periodicities. The results strongly affirm our hypothesis, as the 4-stage GPHT surpasses the 1-stage GPHT (without hierarchical structures), achieving a notable \textbf{2.86\%} reduction in MSE.

\subsubsection{On the Effect of Pretraining} 
\begin{figure}[htbp]
  \centering
  \includegraphics[width=\linewidth]{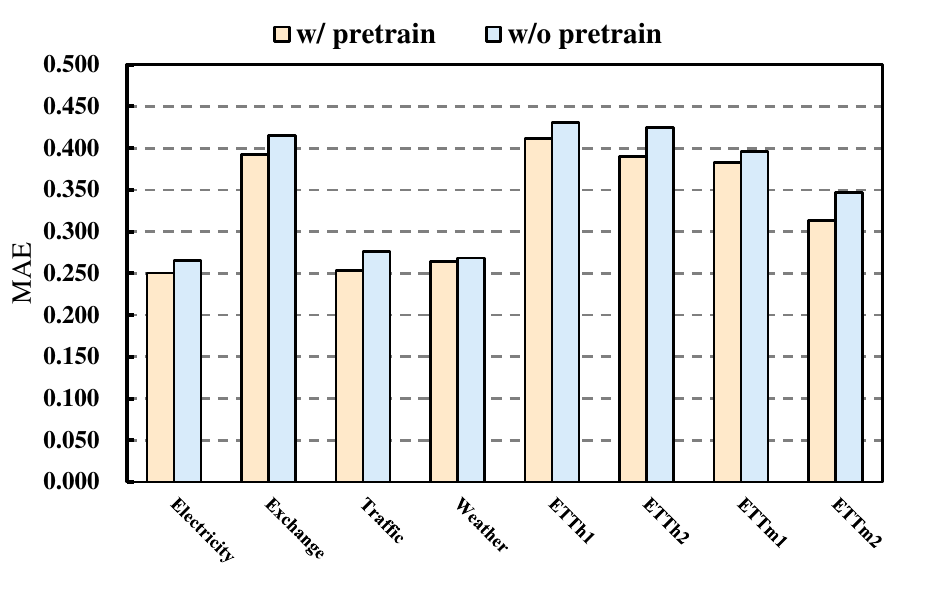}
  \caption{MAE evaluation between GPHT and GPHT without pretraining on benchmark datasets.}
  \label{fig:pretrain}
\end{figure}

In addition to the architecture design, another key aspect is the pretraining procedure on the mixed dataset. Does it pose positive effects on the forecasting performance? We provide quantified results in Figure \ref{fig:pretrain}. Specifically, we compare the performance of fine-tuned GPHT with the one trained from scratch. The MAE evaluations averaged on the horizon length (i.e., $H=96,192,336,720$) are presented. From the results, we can infer that pretraining on the mixed dataset enables the model to leverage commonalities among time series, facilitating better transfer to specific datasets. Compared to GPHT trained from scratch, pretraining results in an average MAE reduction of \textbf{5.75\%}, reaching as high as \textbf{9.65\%} on the ETTm2 dataset.

\section{Conclusion} In this work, we proposed a generative pretrained hierarchical transformer model, namely GPHT, for time series forecasting. It stands out in two key aspects. Conceptually, we explored training a single unified forecasting model that generalizes well across diverse data scenarios and forecasting settings. Technically, we proposed a simple yet effective paradigm that treats time series originating from various scopes as a whole, discarding the heterogeneity and concatenating the values of each variable from different datasets to form the mixed dataset for pretraining. Besides, we replaced conventional one-step generating, which is adopted by most recent forecasting methods, with auto-regressive decoding for better flexibility and performance. We also introduced the hierarchical structure better to capture the diverse patterns in the mixed dataset. We conducted sufficient experiments on 8 widely used datasets in comparison with mainstream self-supervised pretraining models and supervised models, the results demonstrated that GPHT surpasses the baseline models across various fine-tuning and zero/few-shot learning settings in the traditional long-term forecasting task.
% Firstly, it forecasts in an auto-regressive manner under the channel-independent assumption. Consequently, GPHT can be seamlessly adapted to any dataset, including unseen datasets, and forecast at arbitrary horizon lengths. Moreover, it can explicitly model the temporal dependencies in the output series. Secondly, we proposed to conduct pretraining on a mixed dataset comprising various data scenarios. Compared to traditional approaches that only a single dataset with a limited scale is utilized, our proposal could capture the commonalities among time series and therefore facilitate better transfer to specific datasets. 

\begin{acks}
This research was supported by grants from the Joint Research Project of the Science and Technology Innovation Community in Yangtze River Delta (No. 2023CSJZN0200), and the Fundamental Research Funds for the Central Universities. This work also thanked to the support of funding MAI2022C007. 
\end{acks}

%%
%% The next two lines define the bibliography style to be used, and
%% the bibliography file.
\bibliographystyle{ACM-Reference-Format}
\balance
\bibliography{sample-base}

% %%
% %% If your work has an appendix, this is the place to put it.

\appendix

\section{Complexity Comparison}
To better illustrate the proposed GPHT model's computation cost, we provide quantitative results on the Electricity dataset under lookback window $L=336$ and forecasting horizon $H=720$ in Table \ref{tab:complexity}. In summary, the proposed GPHT model is medium-sized compared to the baseline models. Thanks to its straightforward optimization objective, both the pretraining and finetuning processes of GPHT are quite efficient and do not require too much time, especially compared with the pretraining-based approaches. A key drawback of our model might be the inference speed, which is naturally limited by the auto-regressive decoding schema.

% Table generated by Excel2LaTeX from sheet 'Sheet2'
\begin{table*}[tb]
  \centering
  \caption{Computation cost between GPHT and mainstream forecasting approaches.}
    \vspace{-0.1in}
    \begin{tabular}{cc|cccccc}
    \toprule
    \multicolumn{2}{c|}{Methods} & \multicolumn{2}{c}{Params} & \multicolumn{2}{c}{Training Time(per epoch)	} & \multicolumn{2}{c}{Inference Speed(itr/s)} \\
    \midrule
    \multicolumn{2}{c|}{GPHT} & \multicolumn{2}{c}{37.98M(pretraining)/98.50K(finetuning)	} & \multicolumn{2}{c}{20min(pretraining)/254.1s(finetuning)} & \multicolumn{2}{c}{0.34} \\
    \multicolumn{2}{c|}{FPT} & \multicolumn{2}{c}{105.20M(24.00M trainable)} & \multicolumn{2}{c}{3858.8s(finetuning)} & \multicolumn{2}{c}{0.69} \\
    \multicolumn{2}{c|}{SimMTM} & \multicolumn{2}{c}{62.14M(pretraining)/7.76M(finetuning)	} & \multicolumn{2}{c}{73min(pretraining)/946.5s(finetuning)} & \multicolumn{2}{c}{5.98} \\
    \multicolumn{2}{c|}{PatchTST} & \multicolumn{2}{c}{4.27M} & \multicolumn{2}{c}{128.9s} & \multicolumn{2}{c}{9.02} \\
    \multicolumn{2}{c|}{iTransformer} & \multicolumn{2}{c}{5.28M} & \multicolumn{2}{c}{24.7s} & \multicolumn{2}{c}{26.39} \\
    \multicolumn{2}{c|}{TimesNet} & \multicolumn{2}{c}{150.64M} & \multicolumn{2}{c}{1179.6s} & \multicolumn{2}{c}{1.51} \\
    \bottomrule
    \end{tabular}%
    \vspace{-0.1in}
  \label{tab:complexity}%
\end{table*}%

\section{Qualitative Evaluation}
In this section, we provide visualizations of long-term forecasting results to better demonstrate the performance of GPHT and the effectiveness of the hierarchical transformer architecture.

\begin{figure*}[hbp]
  \centering
  \includegraphics[width=\linewidth]{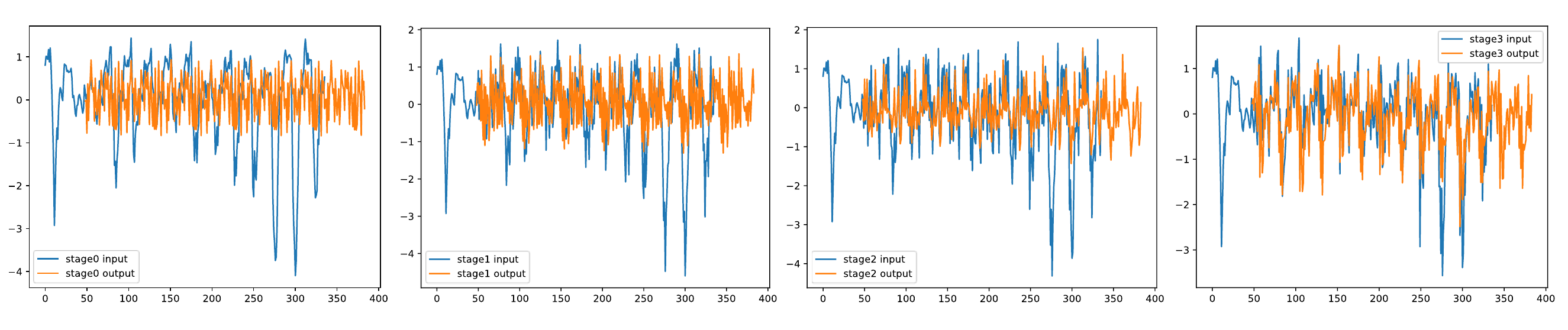}
  \caption{Visualization of the input and corresponding output series of GPHT's multiple stages on a sample from the ETTh1 dataset.}
  \label{fig:hierarchical}
\end{figure*}

\begin{figure*}[hbp]
  \centering
  \includegraphics[width=\linewidth]{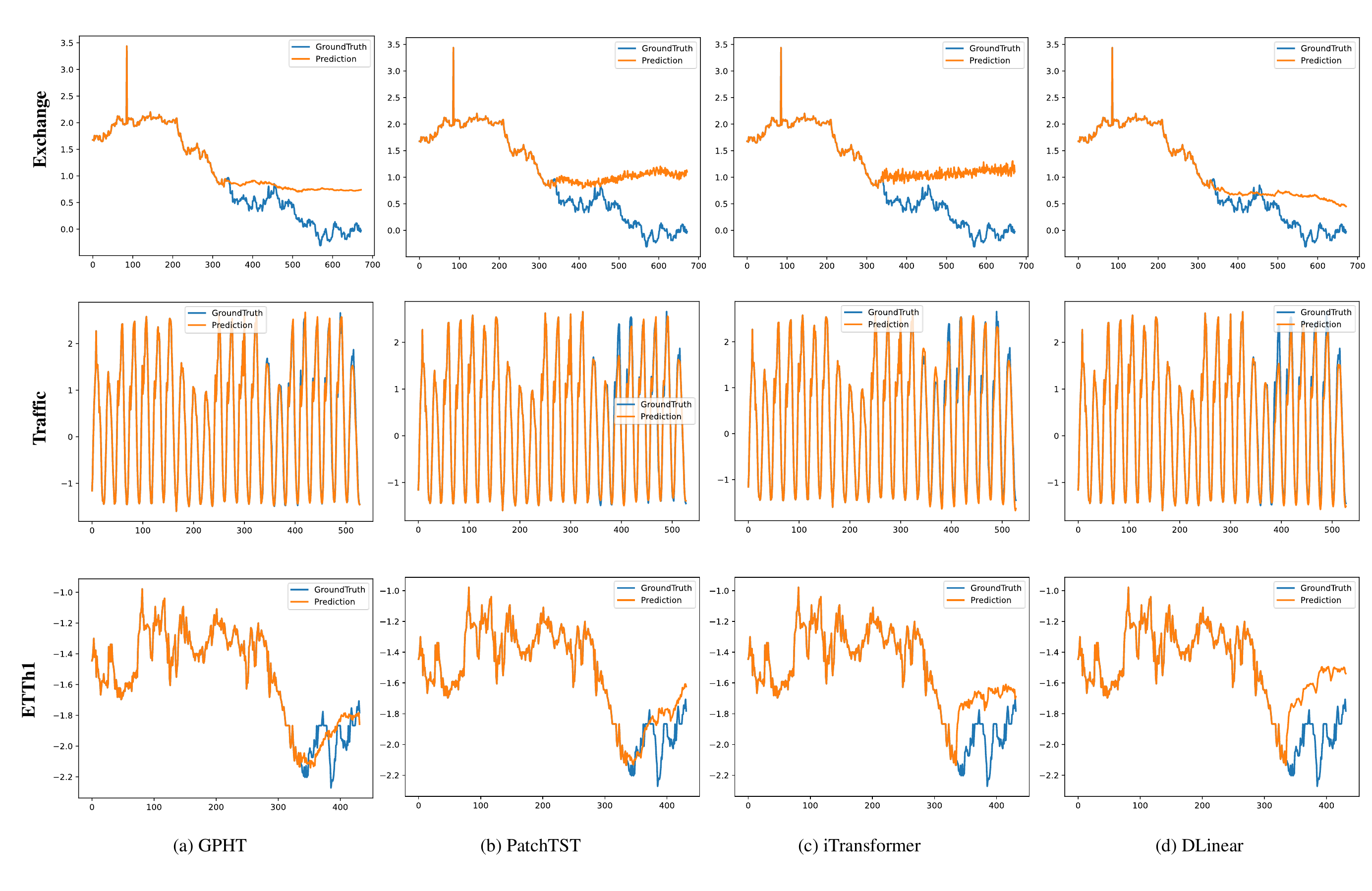}
  \caption{Illustration of forecasting showcases comparing GPHT and baseline models. The lookback window is set to 336 and the forecasting horizon is set to 336, 192, 96 for the Exchange, Traffic, and ETTh1 dataset respectively.}
  \label{fig:visual}
\end{figure*}

We plot a forecasting sample in Figure \ref{fig:hierarchical}, illustrating how GPHT forecasts with the hierarchical architecture and the iterative residual schema. It can be referred that the initial stages predominantly focus on extracting the general periodic patterns from the input series and the latter stages can therefore pay more attention to the specialized trends, since the auto-regressive forecasting results of stage 3 align more closely with the input series, albeit with less periodicity than the preceding stages. The results strongly verify our assumption that the hierarchical architecture can better capture the commonalities and specialties of the mixed pertaining dataset, and the iterative residual schema effectively refines the input for the next stage, eliminating the redundant information in the series. Besides, we provide qualitative comparison between GPHT and mainstream supervised forecasting methods in Figure \ref{fig:visual} on various datasets. Benefiting from the auto-regressive pertaining and the hierarchical architecture, GPHT can better capture the temporal dependencies so as to achieve better performance.

\end{document}